\documentclass[conference,a4paper]{IEEEtran}
\IEEEoverridecommandlockouts

\usepackage{placeins}
\usepackage[hidelinks]{hyperref}
\usepackage[cmex10]{amsmath}
\usepackage{amssymb,amsfonts}
\usepackage{dblfloatfix}

\usepackage[ruled,vlined]{algorithm2e}
\usepackage{graphicx}
\graphicspath{{Figures/PDF/}{Figures/PNG/}}

\usepackage{booktabs}
\usepackage{siunitx}
\usepackage[numbers,compress]{natbib}
\usepackage{texnames}
\usepackage{bm,bbm}
\usepackage{orcidlink}
\begin{document}

\title{\uppercase{An InSAR Phase Unwrapping Framework for Large-scale and Complex Events}
}

 \author{
 \IEEEauthorblockN{
 Yijia~Song$^{1,3}$,
 Juliet~Biggs$^{2}$,
 Alin~Achim$^{1}$,
 Robert~Popescu$^{1,3}$,
 Simon~Orrego$^{2}$,
 Nantheera~Anantrasirichai$^{1,3}$
 }
 \IEEEauthorblockA{
 $^{1}$ Visual Information Laboratory, School of Computer Science, University of Bristol, UK\\
 $^{2}$ COMET, School of Earth Sciences, University of Bristol, UK\\
 $^{3}$ COMET, School of Computer Science, University of Bristol, UK
 }
 }

\maketitle
\begin{abstract}
	Phase unwrapping remains a critical and challenging problem in InSAR processing, particularly in scenarios involving complex deformation patterns. In earthquake-related deformation, shallow sources can generate surface-breaking faults and abrupt displacement discontinuities, which severely disrupt phase continuity and often cause conventional unwrapping algorithms to fail. Another limitation of existing learning-based unwrapping methods is their reliance on fixed and relatively small input sizes, while real InSAR interferograms are typically large-scale and spatially heterogeneous. This mismatch restricts the applicability of many neural network approaches to real-world data. In this work, we present a phase unwrapping framework based on a diffusion model, developed to process large-scale interferograms and to address phase discontinuities caused by deformation. By leveraging a diffusion model architecture, the proposed method can recover physically consistent unwrapped phase fields even in the presence of fault-related phase jumps. Experimental results on both synthetic and real datasets demonstrate that the method effectively addresses discontinuities associated with near-surface deformation and scales well to large InSAR images, offering a practical alternative to manual unwrapping in challenging scenarios.
\end{abstract}


\section{Introduction}
\label{sec:intro}

Interferometric Synthetic Aperture Radar (InSAR) provides an effective means for observing surface deformation associated with earthquakes, volcanic activity, and other geophysical phenomena~\cite{Poland2022Volcano, Ebmeier2018Synthesis, popescu2025unsupervised}. A fundamental step in InSAR processing is phase unwrapping, which aims to recover the absolute interferometric phase from measurements that are inherently wrapped modulo $2\pi$. While the wrapped phase is directly observed, the unwrapped phase encodes the true deformation signal and serves as the basis for subsequent geophysical interpretation. Recovering this quantity requires resolving both the continuous phase field and the unknown integer ambiguities, rendering phase unwrapping an ill-posed inverse problem. 
Although a wide range of algorithms have been developed, reliable unwrapping remains difficult in practice, particularly in the presence of noise, decorrelation, or sharp spatial variations in deformation~\cite{Yan2025A}.

Existing phase unwrapping techniques are commonly categorized into path-based and optimization-based methods. Path-based approaches, such as branch-cut algorithms~\cite{goldstein1998radar}, unwrap the phase by integrating along selected paths while avoiding inconsistencies introduced by residues. In contrast, optimization-based methods formulate phase unwrapping as a global consistency problem, with the Statistical-cost, Network-flow Algorithm for Phase Unwrapping (\textsc{SNAPHU})~\cite{chen2002phase} being a representative example. By solving a network-flow–based optimization problem, these methods enforce consistency across the entire interferogram. Despite their effectiveness under favorable conditions, both categories tend to degrade in regions of low coherence or in scenarios involving large deformation gradients, where local errors can propagate and compromise the final solution.

Recent deep learning techniques have been explored for InSAR phase unwrapping, although task-specific designs remain relatively limited. 
PhaseNet~\cite{phasenet} reframed unwrapping as a wrap-count classification problem, and subsequent work PhaseNet~2.0~\cite{phasenet2} refined the network architecture and training strategy to improve performance. Alternative formulations include unsupervised and physics-inspired approaches, e.g. U3Net~\cite{U3NET}, to incorporate coherence information and reconstruction constraints through deep unrolling. SQD-LSTM~\cite{SQD} models the wrapped-to-unwrapped relationship using recurrent structures. The models originally developed for image restoration have also been adapted to this task.  Restormer~\cite{Restormer} has been applied by tailoring loss functions to phase data~\cite{U3NET}. PU-GAN~\cite{Zhou2022PU-GAN} employs conditional enerative
adversarial network. Diffusion probabilistic models~\cite{ho2020ddpm,nichol2021improved} have recently demonstrated strong performance in image reconstruction tasks, and UnwrapDiff~\cite{Song2026UnwrapDiff} has shown superior performance compared with the aforementioned methods. 

Existing methods, however, tend to perform well only on relatively simple deformation scenarios. Consequently, this work focuses on two practical aspects that are frequently encountered in real InSAR phase unwrapping tasks. First, the proposed framework is designed to accommodate large-scale interferograms, where spatial heterogeneity, long-range phase variations, and non-uniform data quality are commonly observed. These characteristics challenge methods that rely on fixed-size inputs or strictly local processing assumptions, and motivate a formulation that can operate consistently across extended spatial domains.

Second, particular attention is given to deformation patterns associated with shallow earthquakes, in which surface-breaking faults introduce abrupt and spatially localized phase discontinuities across fracture boundaries. Such discontinuities violate the continuity assumptions underlying many conventional and learning-based unwrapping methods, often resulting in localized failures or error propagation near fault zones. By explicitly accounting for both large spatial extents and deformation-induced phase jumps within a unified framework, this work aims to provide a systematic treatment of interferograms exhibiting complex spatial structures and non-smooth deformation behavior. The proposed framework is described in detail in the following section, and its performance is examined using both synthetic data and real earthquake interferograms.



\begin{figure}[t]
    \centering
    \includegraphics[width=\columnwidth]{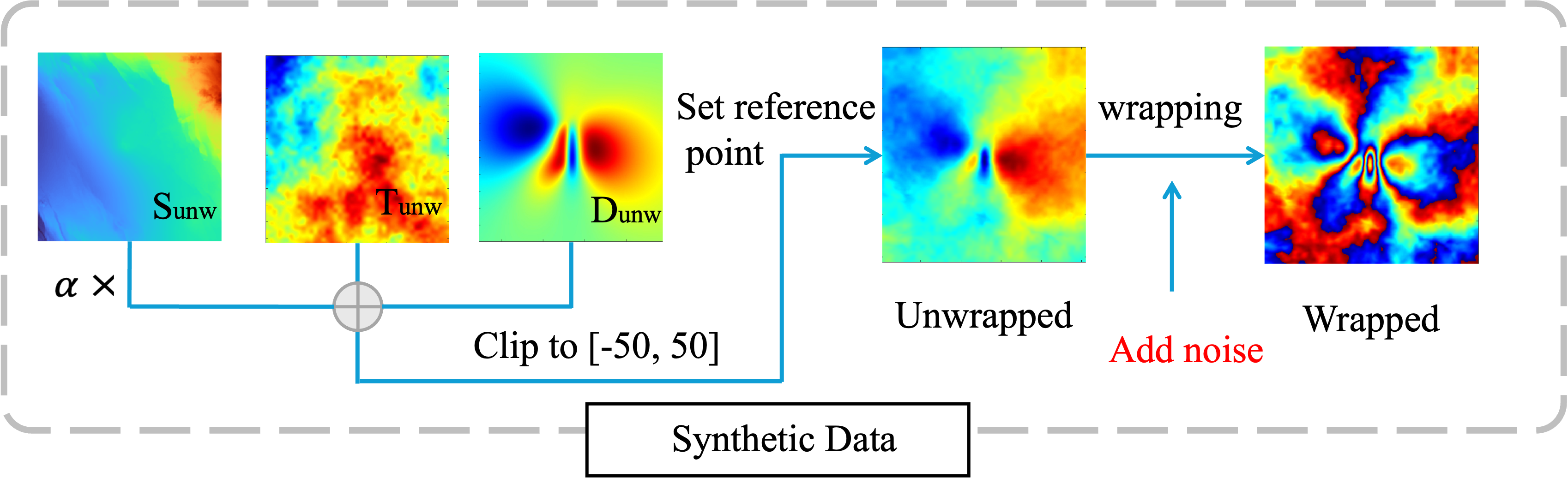}
    \caption{Overview of the Synthetic Data Generation Pipeline.}
    \label{fig:synthetic}
\end{figure}

\section{Synthetic InSAR data}
\label{sec:syndata}

\subsection{Phase Unwrapping Formulation}
InSAR phase unwrapping aims to infer a spatially continuous phase field from interferometric observations that are inherently wrapped within a principal interval. Given a wrapped interferogram $\phi_w(x)\in(-\pi,\pi]$, the underlying objective is to estimate the corresponding unwrapped phase $\phi(x)$ that explains the observed measurements. The wrapping operation can be written as
\begin{equation}
\phi_w(x) = \mathrm{mod}\big(\phi(x)+\pi,\,2\pi\big)-\pi,
\end{equation}
where $\phi(x)$ represents the unknown continuous phase signal. In practical scenarios, only $\phi_w(x)$ is available, and the unwrapping task consists of resolving the unknown $2\pi$ ambiguities at each spatial location.
Moreover, real interferograms contain measurement noise, decorrelation, discontinuities across the $(-\pi,\pi]$ boundary, and missing or unreliable pixels. 

\subsection{Synthetic interferograms}
The synthetic dataset used in this study is constructed to emulate the dominant physical and atmospheric contributions commonly observed in earthquake-related InSAR interferograms. An overview of the synthetic data generation pipeline is illustrated in Fig.~\ref{fig:synthetic}.
The synthetic unwrapped interferograms are generated by explicitly modelling three principal components: coseismic surface deformation, stratified atmospheric delays, and turbulent atmospheric artefacts, following established InSAR simulation practices~\cite{anantrasirichai2019deep}. 
These components are combined linearly to form
\begin{equation}
\phi_{\text{sim}} = D + S + T + \phi_{\text{noise}},
\end{equation}
where $D$, $S$, and $T$ denote deformation, stratified atmosphere, and turbulent atmosphere, respectively. $\phi_{\text{noise}}$ includes decorrelation and measurement noise.


\vspace{2mm}
\noindent\textbf{Deformation Generation.}



Surface deformation ($D$) is generated from earthquake sources using the Okada elastic dislocation model~\cite{Okada1985}, with a focus on shallow ruptures. Source parameters, including location, depth, strike, dip, rake, and slip, are sampled using a Monte Carlo strategy to produce diverse deformation patterns. The resulting surface displacement is projected into the radar line-of-sight to obtain the deformation-induced phase,
\begin{equation}
\phi_{\text{def}}(x,y) = \frac{4\pi}{\lambda}\, \big(\mathcal{O}(x,y \mid \boldsymbol{\theta}) \cdot \mathbf{l}\big),
\end{equation}
where $\mathcal{O}(\cdot)$ denotes the Okada operator and $\mathbf{l}$ is the line-of-sight unit vector.

To explicitly model deformation discontinuities associated with surface-breaking faults, random surface fractures are introduced for a subset of shallow events. Across a fracture line $\Gamma$, the deformation phase is locally modified as
\begin{equation}
\phi_{\text{def}}'(x,y) = \phi_{\text{def}}(x,y) + \Delta \phi_{\text{def}}\, \mathrm{sign}\!\big((x,y)\cdot\mathbf{n}\big),
\end{equation}
where $\mathbf{n}$ is the normal vector of the fracture and $\Delta \phi_{\text{def}}$ denotes a phase offset applied with opposite signs on the two sides of the fracture. 
To the best of our knowledge, this simple fracture-based modification has not been explicitly incorporated in existing synthetic InSAR phase unwrapping datasets, and is introduced here to emulate abrupt phase jumps observed near surface-breaking faults.

In realistic earthquake scenarios, surface deformation may be influenced by more than one seismic source within the same region, resulting in overlapping deformation patterns. To account for this additional complexity, a subset of the training data is generated using two earthquake sources within a single scene. In these cases, deformation fields from the two sources interact spatially, producing more complex phase gradients and discontinuities than those associated with a single event. By including such multi-source samples in the dataset, the training data better reflects the variability and complexity observed in real interferograms, and helps reduce reliance on overly simplified deformation assumptions. These samples are processed in the same manner as single-source cases in subsequent steps.

\begin{figure*}[t]
    \centering
    \includegraphics[width=\textwidth]{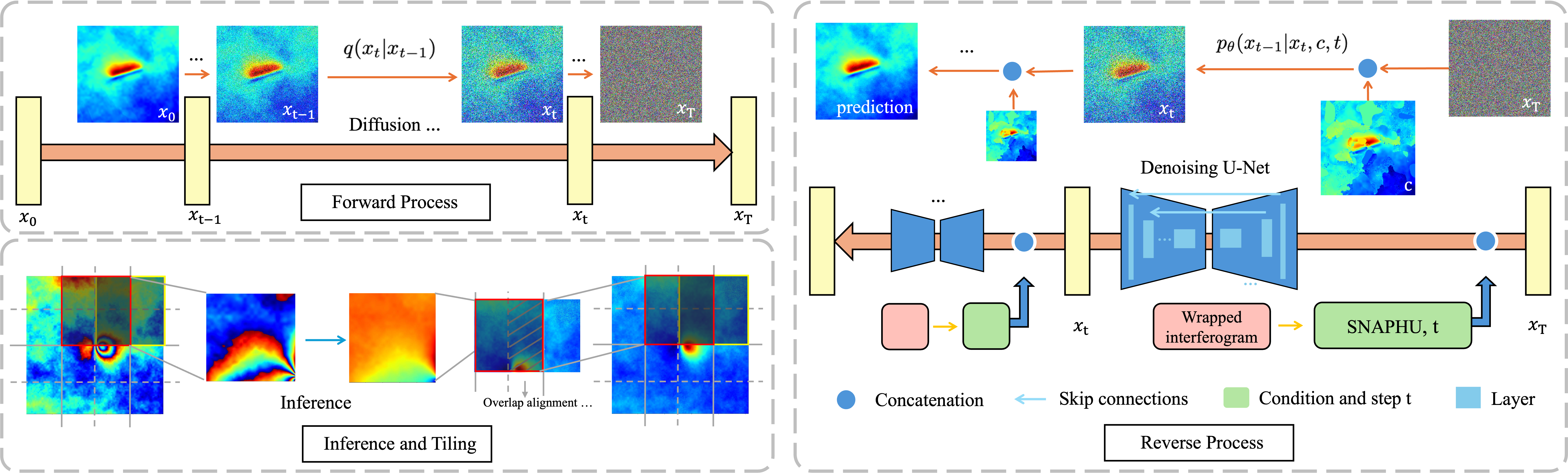}
    \caption{Diagram of the proposed INSAR unwrapping framework.}
    \label{fig:diffusion}
\end{figure*}

\vspace{2mm}
\noindent\textbf{Atmosphere Generation.} 
Stratified atmospheric delays ($S$) are simulated using products from the Generic Atmospheric Correction Online Service (GACOS)~\cite{GACOS}, which derives zenith total delay maps from numerical weather prediction data. These delays are interpolated to the interferogram grid and mapped into the radar line-of-sight to represent elevation-correlated atmospheric effects.
Turbulent atmospheric artefacts ($T$) are generated using a statistical model with an exponentially decaying spatial covariance~\cite{Biggs2007AtmosphericNoise}, consistent with observations from large-scale Sentinel-1 interferogram archives. Synthetic turbulence realisations are produced via covariance matrix sampling and Cholesky decomposition to preserve realistic spatial correlations.


\vspace{2mm}
\noindent\textbf{Noise modelling.}
To improve the realism of the synthetic data and enhance model robustness, patchy noise is introduced to simulate spatially structured phase inconsistencies commonly observed in real InSAR interferograms. Patchy artefacts are generated by thresholding a smoothed Gaussian random field, producing irregular regions with locally inconsistent phase behaviour. Within each patch, phase values are perturbed by random offsets in the range \(0\!-\!2\pi\), representing localized phase errors that are difficult to model explicitly.

\section{Methodology}
\label{sec:Methodology}


An overview of the proposed framework is shown in Fig.~\ref{fig:diffusion}, consisting of a diffusion-based training stage and a tiling-based inference stage, where large interferograms are decomposed into overlapping tiles, unwrapped independently, and fused to reconstruct the full-resolution phase.

\subsection{Diffusion-based Phase Unwrapping}

The proposed approach formulates InSAR phase unwrapping within a conditional diffusion framework. Model training follows the diffusion probabilistic formulation, while inference is performed using DDIM~\cite{Song2021DDIM} for efficient sampling.The phase estimate obtained from SNAPHU is provided to the network as a conditioning input, supplying large-scale structural information. Guided by this prior, the diffusion process progressively refines the phase field and corrects localized inconsistencies that arise in regions affected by noise or steep phase variations.

\vspace{2mm}
\noindent\textbf{Forward Diffusion Process}: 
Let $\phi$ denote the clean unwrapped phase field, and define $x_0=\phi$. The forward diffusion process gradually perturbs $x_0$ into a noisy variable $x_T$ through a Markov chain:
\begin{equation}
q(x_t \mid x_{t-1}) = \mathcal{N}\!\left(x_t; \sqrt{1-\beta_t}\,x_{t-1}, \beta_t \mathbf{I}\right),
\end{equation}
where $\{\beta_t\}_{t=1}^T$ follows a cosine variance schedule. After $T$ diffusion steps, the distribution of $x_T$ approaches an isotropic Gaussian.

\vspace{2mm}
\noindent\textbf{Conditional Reverse Process}: 
The reverse process aims to reconstruct the clean phase field from noisy observations. We employ the conditioning generated using  \textsc{SNAPHU}~\cite{chen2002phase}. Note that this approach is not limited to \textsc{SNAPHU}; however, \textsc{SNAPHU} has been widely used and has proven effective in general.
This SNAPHU conditioning denoted as $c$ is concatenated with $x_t$ as an additional input channel to the U-Net backbone. The network is trained to predict the noise term $\epsilon_\theta(x_t, c, t)$, enabling it to leverage the global prior from \textsc{SNAPHU} while correcting its local and systematic errors.

\vspace{2mm}
\noindent\textbf{DDIM Sampling}: 
During inference, we employ Denoising Diffusion Implicit Models (DDIM) for accelerated sampling. Let $\alpha_t = \prod_{s=1}^t (1-\beta_s)$. Given a noisy sample $x_t$, the clean estimate $\hat{x}_0$ is first computed as:
\begin{equation}
\hat{x}_0 = \frac{x_t - \sqrt{1-\alpha_t}\,\epsilon_\theta(x_t, c, t)}{\sqrt{\alpha_t}}.
\end{equation}

The sample at the previous timestep is then obtained via:
\begin{equation}
x_{t-1} = \sqrt{\alpha_{t-1}}\,\hat{x}_0
        + \sqrt{1-\alpha_{t-1}-\sigma_t^2}\,\epsilon_\theta(x_t, c, t)
        + \sigma_t z,
\end{equation}
where $z \sim \mathcal{N}(0,\mathbf{I})$. The variance term $\sigma_t$ is controlled by a parameter $\eta \in [0,1]$:
\begin{equation}
\sigma_t = \eta 
\sqrt{\frac{1-\alpha_{t-1}}{1-\alpha_t}}
\sqrt{1-\frac{\alpha_t}{\alpha_{t-1}}}.
\end{equation}
When $\eta=0$, the sampling process becomes deterministic. To further reduce inference cost, sampling is performed over a subsequence of timesteps $\{t_k\}_{k=1}^{\tau}$, where $\tau \ll T$.

\vspace{2mm}
\noindent\textbf{Training and Loss Function}:
Following the improved diffusion training objective~\cite{nichol2021improved}, the network is trained to predict the injected noise at each diffusion step using a weighted mean squared error:
\begin{equation}
\mathcal{L}_{\text{diff}} =
\mathbb{E}_{t, x_0, \epsilon}\!\left[
w_t \cdot \left\| \epsilon - \epsilon_\theta(x_t, c, t) \right\|_2^2
\right],
\end{equation}
where $w_t$ is a variance-schedule-dependent weighting factor. The same noise predictor is reused during DDIM-based inference, enabling efficient generation without modifying the training objective.

\subsection{Large-Scale Inference via Overlapping Tiling}

Real-world InSAR interferograms are typically large-scale and spatially heterogeneous, making direct inference on full-resolution images impractical due to memory and computational constraints. To enable scalable inference while preserving spatial continuity, we adopt an overlapping tiling strategy during the inference stage.

Specifically, a large interferogram $\mathbf{X}\in\mathbb{R}^{H\times W}$ is decomposed into a set of partially overlapping tiles $\{\mathbf{X}_k\}_{k=1}^{K}$ of fixed size $256\times 256$. Each tile is independently processed by the trained diffusion model using DDIM sampling, producing a corresponding unwrapped phase estimate $\{\hat{\mathbf{Y}}_k\}_{k=1}^{K}$. The overlap between adjacent tiles is introduced to mitigate boundary artifacts and to improve consistency across tile borders.

Let $\Omega_k$ denote the spatial support of the $k$-th tile in the full image domain. The final unwrapped phase $\hat{\mathbf{Y}}$ is obtained by aggregating all tile-wise predictions through weighted averaging:
\begin{equation}
\hat{\mathbf{Y}}(x) = 
\frac{\sum_{k:\,x\in\Omega_k} w_k(x)\,\hat{\mathbf{Y}}_k(x)}
     {\sum_{k:\,x\in\Omega_k} w_k(x)},
\end{equation}
where $w_k(x)$ is a spatial weighting function that assigns lower weights to pixels near tile boundaries. In this work, $w_k(x)$ is chosen as a smooth window function to emphasize the central region of each tile and reduce edge-related inconsistencies. This overlapping tiling formulation allows the proposed framework to process interferograms of arbitrary size while maintaining local coherence across tile boundaries. By combining tile-wise diffusion-based refinement with spatially weighted aggregation, the method achieves scalable inference without modifying the network architecture or the training objective.

\section{Experimental results}
\label{sec:Experiment}

All experiments use a consistent parameter setting. Inference is performed using overlapping tiles of size $256 \times 256$ with a 128-pixel overlap and DDIM sampling with 50 steps. For synthetic data, noise levels are randomly varied between 5\% and 30\%. The dataset consists of
11,000 samples in total, with 10,000 used for training and
1,000 reserved for testing.

To examine the effect of different inference strategies on large-scale phase unwrapping, we compare two approaches applied during the inference stage. In the first approach, the entire wrapped interferogram is resized to match the network input size and processed in a single forward pass. In the second approach, the wrapped interferogram is decomposed into overlapping tiles, which are unwrapped individually and then merged to reconstruct the full-resolution result.


\begin{figure}[t]
    \centering
    \includegraphics[width=\columnwidth]{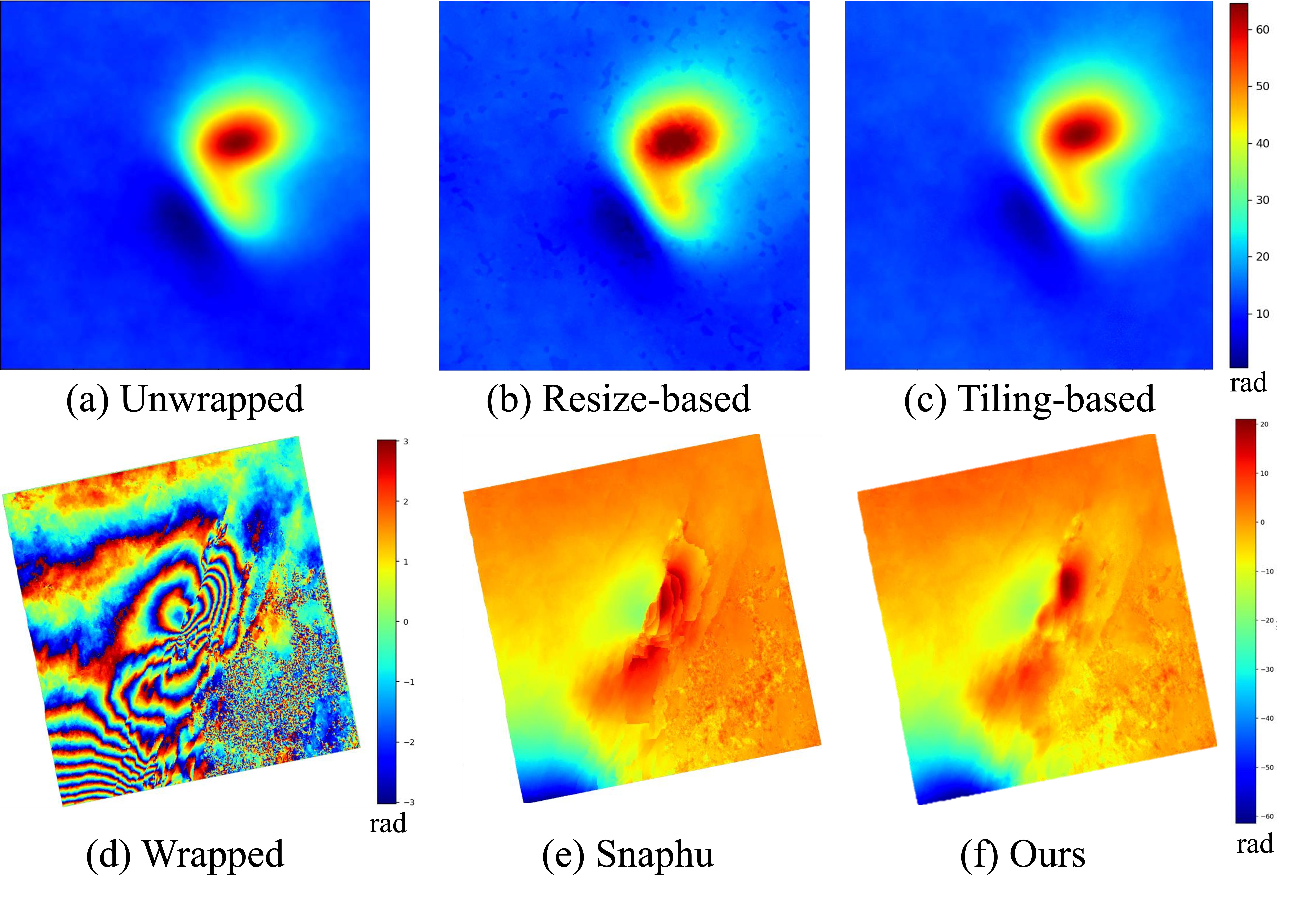}
    \caption{Qualitative comparison of phase unwrapping results on synthetic and real data shown on top and bottom rows, respectively.}
    \label{fig:result}
\end{figure}

The comparison between Fig.\ref{fig:result}(b) and Fig.\ref{fig:result}(c) indicates that the resize-based inference strategy is less effective in suppressing noise, with larger errors observed around regions of abrupt wrapped phase jumps, as shown in the corresponding error maps. In contrast, the proposed tiling-based strategy better preserves local phase structures and achieves improved reconstruction in noisy and discontinuous areas. Some variation in reconstruction accuracy across different tiles is still observed. Quantitatively, the tiling-based approach achieves a lower Normalized Root Mean Square Error (NRMSE), defined as the reconstruction error normalized by the dynamic range of the reference unwrapped phase, reducing the error from 0.68\% (resize-based) to 0.46\%.

For the real-data experiment in Fig.~\ref{fig:result}(d-e), we use an ascending Sentinel-1 interferogram spanning 5 December 2024 to 10 January 2025. The interferogram exhibits a central phase discontinuity, additional deformation in the lower-left region, and strong noise in the lower-right area. The proposed method is qualitatively compared with SNAPHU, as ground-truth unwrapped phase is not available for real data.

On the left side of the central fracture, both methods produce generally consistent unwrapped phase. However, on the right side, the SNAPHU result exhibits noticeable step-like phase jumps, while the proposed method yields a more continuous phase field. The deformation associated with the secondary source in the lower-left region is recovered by both methods. In the lower-right noisy area, SNAPHU shows limited suppression of noise-related artefacts, whereas the proposed method produces a comparatively smoother result, although residual noise remains.


\section{Conclusion}

This paper presents a diffusion-based framework for InSAR phase unwrapping, addressing two practical challenges in real applications: large-scale interferograms and deformation-induced phase discontinuities. By combining a conditional diffusion model with a tiling-based inference strategy, the proposed method enables efficient processing of large interferograms while maintaining phase continuity in high-gradient regions.
Experiments on synthetic data demonstrate that tiling-based inference is more robust than resize-based processing, particularly near phase jumps. Qualitative results on real interferograms indicate improved spatial consistency compared with conventional methods in areas affected by surface fractures and strong noise. 


\small
\bibliographystyle{IEEEtranN}
\bibliography{references}

\end{document}